\documentclass[letterpaper, 10 pt, conference]{ieeeconf}  

\IEEEoverridecommandlockouts                              

\usepackage{amsmath} 
\usepackage{amssymb}  
\usepackage{cite}
\usepackage{booktabs}
\usepackage[subnum]{cases}
\usepackage[subrefformat=parens,labelformat=parens]{subfig}\usepackage{balance}
\usepackage{balance}
\usepackage{color}
\usepackage{tabularx}
\usepackage{graphicx,dblfloatfix}
\usepackage{listings}
\usepackage{textcomp}
\usepackage{url}
\usepackage{multirow}
\usepackage{ragged2e}
\newcolumntype{Y}{>{\RaggedRight\arraybackslash}X}

\usepackage{bm}
\usepackage[colorinlistoftodos]{todonotes}
\usepackage{hyperref}
\usepackage[normalem]{ulem}

\newcommand{\bfq}{\mathbf{q}}
\newcommand{\bfx}{\mathbf{x}}
\newcommand{\bfu}{\mathbf{u}}
\newcommand{\bfw}{\mathbf{w}}

\newcommand{\bfz}{\mathbf{z}}
\newcommand{\bfT}{\mathbf{T}}
\newcommand{\bfA}{\mathbf{A}}
\newcommand{\bfB}{\mathbf{B}}
\newcommand{\bfS}{\mathbf{S}}
\newcommand{\bfU}{\mathbf{U}}
\newcommand{\bfX}{\mathbf{X}}
\newcommand{\bfQ}{\mathbf{Q}}
\newcommand{\bfR}{\mathbf{R}}
\newcommand{\bfE}{\mathbf{E}}
\newcommand{\bfG}{\mathbf{G}}
\newcommand{\bfH}{\mathbf{H}}

\usepackage[printonlyused]{acronym}

\title{Printing-while-moving: a new paradigm for large-scale robotic 3D Printing} 

\author{Mehmet Efe Tiryaki, Xu Zhang, Quang-Cuong Pham
  \thanks{The authors are with Singapore Centre for 3D Printing
    (SC3DP), School of Mechanical and Aerospace Engineering, Nanyang
    Technological University, Singapore. Email: cuong@ntu.edu.sg.} }
\begin{document}

\maketitle
\thispagestyle{empty}
\pagestyle{empty}

\begin{abstract}
  Building and Construction have recently become an exciting
  application ground for robotics. In particular, rapid progress in
  materials formulation and in robotics technology has made robotic 3D
  Printing of concrete a promising technique for in-situ
  construction. Yet, scalability remains an important hurdle to
  widespread adoption: the printing systems (gantry-based or
  arm-based) are often much larger than the structure to be printed,
  hence cumbersome. Recently, a mobile printing system -- a
  manipulator mounted on a mobile base -- was proposed to alleviate
  this issue: such a system, by moving its base, can potentially print
  a structure larger than itself. However, the proposed system could
  only print while being stationary, imposing thereby a limit on the
  size of structures that can be printed in a single take. Here, we
  develop a system that implements the printing-while-moving paradigm,
  which enables printing single-piece structures of arbitrary sizes
  with a single robot. This development requires solving motion
  planning, localization, and motion control problems that are
  specific to mobile 3D Printing. We report our framework to address
  those problems, and demonstrate, for the first time, a
  printing-while-moving experiment, wherein a 210\,cm $\times$ 45\,cm
  $\times$ 10\,cm concrete structure is printed by a robot arm that
  has a reach of 87\,cm.
\end{abstract}


\section{Introduction}

Digital fabrication in construction has become one of the main foci of
robotic research in recent years, with the perspective of fully
autonomous construction. While various in-situ construction approaches
have been demonstrated for steel-frame building and brick wall
construction
~\cite{Giftthaler2017,RILEMTechLett,Dörfler2016,Lussi2018}, there is
also active research in concrete robotic 3D Printing, which we review
in detail in Section~\ref{sec:related_work}.

Despite rapid developments in printable material
formulation~\cite{le2012hardened,zareiyan2017interlayer,tay2016processing}
and system design, scalability remains a major hurdle to the
widespread adoption of concrete 3D printing in B\&C.  For most of the
existing gantry-based and arm-based printing systems, the sizes of the
printed structures are constrained either by the limited volume of the
gantry, or by the reach of the robot arm. To alleviate the scalability
issue, some mobile printing systems have been demonstrated, where a
robot manipulator is mounted on a mobile platform to perform the
printing~\cite{zhang2018large}. However, up to now, the printing can
only be executed when the printer is stationary, which still places
limitations on the size of structures that can be printed \emph{in a
  single take}.

\begin{figure}[t]
    \centering
    \includegraphics[width=\columnwidth]{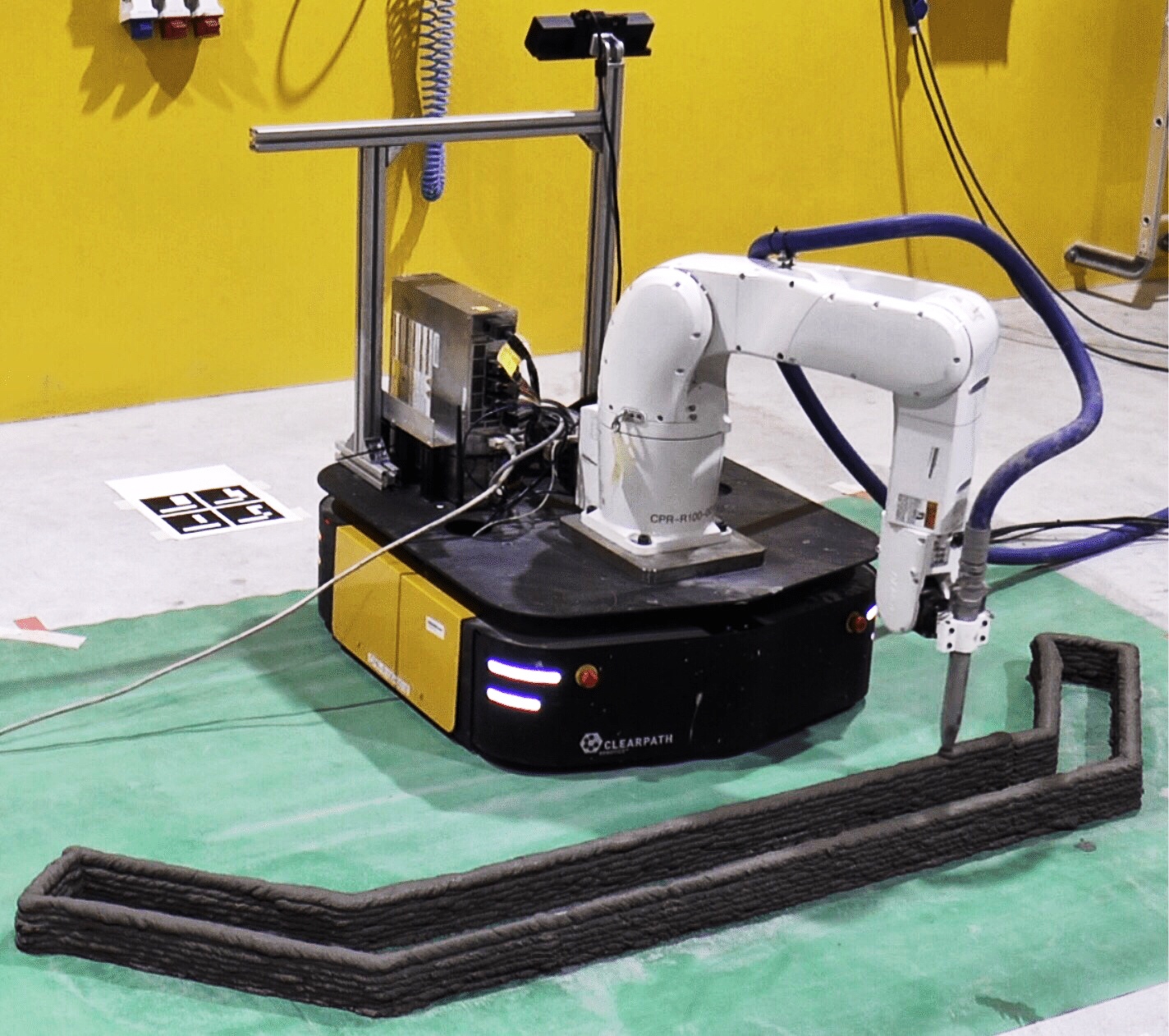}
    \caption{The mobile robotic 3D Printing system presented in this
      work. A video of the printing experiment is available at
      \url{https://youtu.be/ZIbY00iTFwY}.}
    \label{fig:robot}
\end{figure}

In this paper, we propose a 3D printing system that implements
printing-while-moving paradigm. This paradigm enables printing
single-piece structures of arbitrary sizes with a single mobile robot
printer. We demonstrate, \emph{for the first time}, the actual
printing of a single-piece concrete structure by a mobile robot
printer. The size of the structure is 210\,cm $\times$ 45\,cm $\times$
10\,cm (length, width, height), which is larger than the reach of the
robot arm (87\,cm).

Achieving printing-while-moving requires addressing several
challenges. The motions of the robot manipulator and the mobile
platform must be carefully planned and coordinated. In addition,
precise robot localization and feedback motion control are necessary
to ensure that the nozzle deposits concrete at the right place and
with the right speed. This is particularly important in the
layer-by-layer process at hand: nozzle position offsets of more than
1\,cm between two consecutive layers can cause the structure to
collapse.

The remainder of this article is organized as follows. In
Section~\ref{sec:related_work}, we review existing 3D Printing systems
for B\&C. In Section~\ref{sec:system_development}, we present in
detail the proposed mobile printing system. In
Section~\ref{sec:experiments}, we report the results of the accuracy
and precision assessment, as well as the actual printing experiment.
In Section~\ref{sec:conclusion}, we conclude by discussing the
limitations of our current system and sketch some directions for
future work.

\section{Related work}
\label{sec:related_work}

Many groups work on automatizing construction process using 3D
printing techniques and suggested different methods. One of the widely
adopted methods is gantry-based concrete 3D printers
\cite{khoshnevis2006mega,lim2012developments,tay20173d}. Due to the
resemblance to the conventional additive manufacturing methods with
printing nozzle moving in the Cartesian space, it is relatively easier
to design the printer itself, and it is also straightforward to slice
the printed shape and plan the motion of the nozzle. However, the
printing area of these systems is limited to the space enclosed by the
gantry. Thus, it is not possible to print any construction wider than the
foothold distance of the gantry. Moreover, these system are in general
extremely heavy and requires pre-installation on the printing site
before each printing operation.

Another class of widely used 3D concrete printing systems is the arm-based
systems \cite{gosselin2016large,Keating2017}. These systems offer more
flexible printing experience as they can orient their nozzle in 3
dimensional rotational space. Therefore, these systems can potentially print
more complex contours. In addition, installation process of these
system on the printing site is relatively easier than the gantry-based
systems. However, they can still only print inside of the reachable
space of the arm, and this space is further limited with the
orientation of the nozzle and kinematic singularities at some parts of
the reachable space.

In order to resolve scalability problem of the arm-based approaches,
there are some groups proposing arm-based printing systems on a mobile
base. Digital Construction Platform (DCP) by MIT \cite{Keating2017} is
one of the examples of such systems. The DCP is composed of a robot
arm, which can reach $10m$ radially and $14m$ vertically, and a track
based nonholonomic mobile platform. This system uses
printing-upon-arrival strategy, hence it works in a quasi-static
manner. Once it moves into the printing site, it stabilizes itself to
the ground and prints later. This approach does not only limit the
size of structure to be printed, but also limits the shape of the
structure, since at the end of the printing mobile printer should be
able to leave the printing site. Another example of mobile concrete 3D
printers is our previous work \cite{zhang2018large}, in which multiple 
arm-based printers with holonomic mobile base are used to
enlarge printable area. This printing system is also based on
printing-upon-arrival strategy.

Another notable printing approach for 3D concrete printing is
Minibuilders \cite{Minibuilders}, which is composed of three small mobile
robots. The first robot prepares a short concrete wall by following
pre-installed strips. The second robot is placed on the top of the
wall and continues building it while moving through the contour of the
wall. The last robot performs surface finishing of the wall.

\section{System development}
\label{sec:system_development}

\subsection{Mobile printing system}
\label{sec:mechanical_design}

\begin{figure}[htp]
    \centering
    \includegraphics[width=\columnwidth]{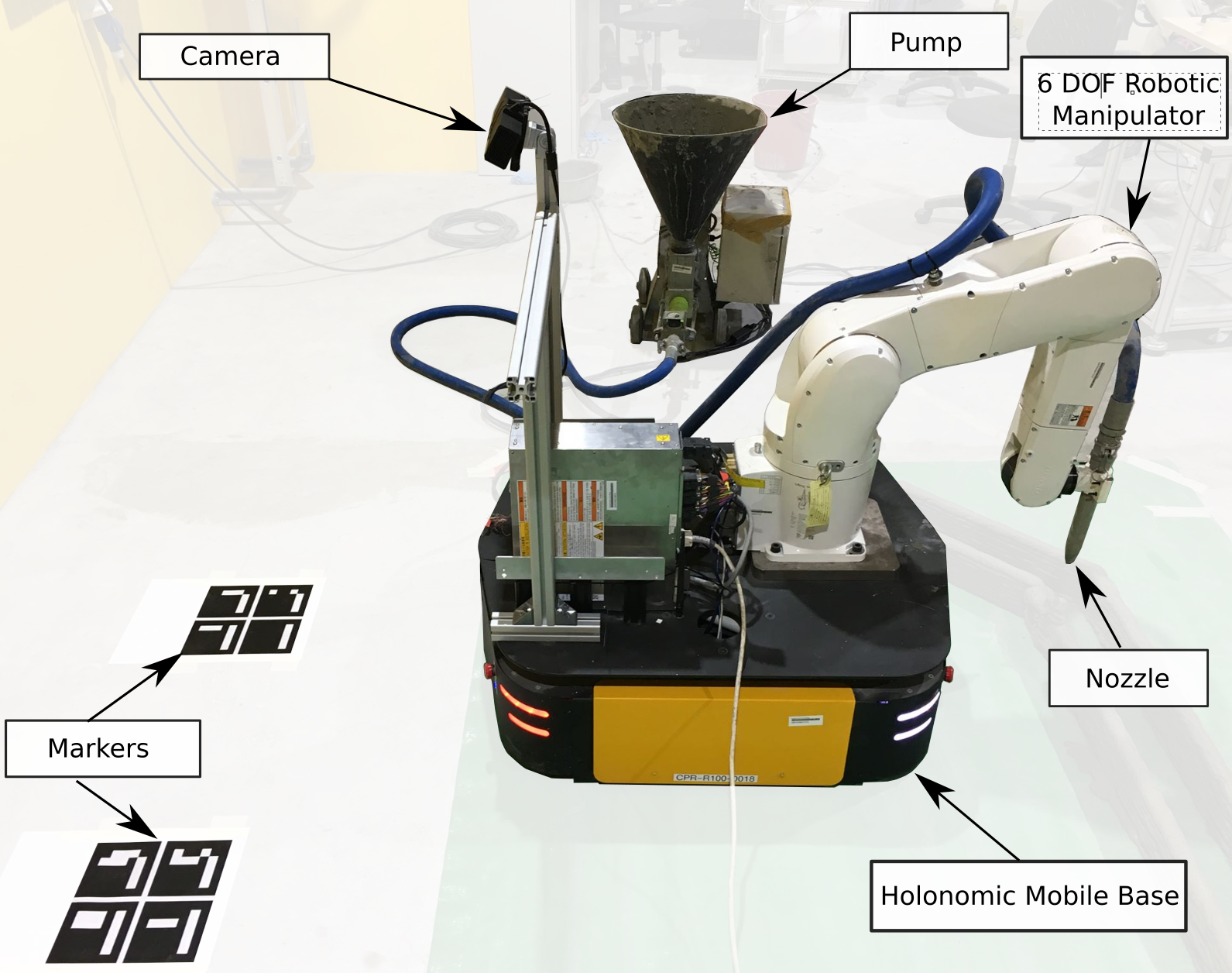}
    \includegraphics[width=\columnwidth]{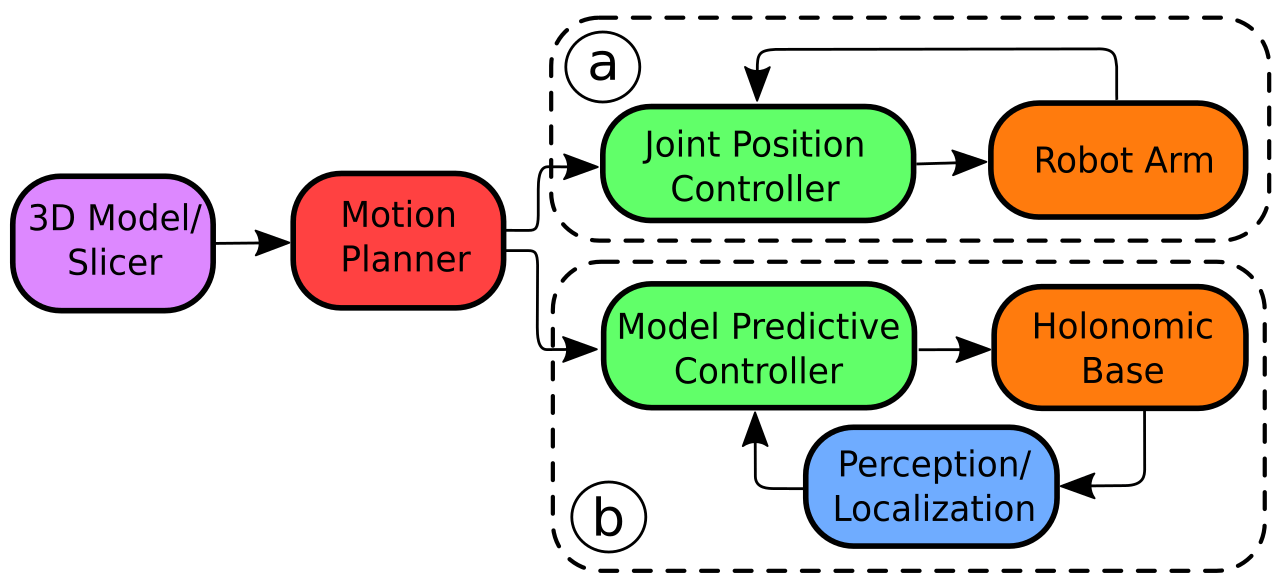}
    \caption{\textbf{Top}: Hardware setup, comprising: holonomic
      mobile base, 6-DoF robot manipulator, camera, markers, pump,
      print nozzle. \textbf{Bottom}: Pipeline of the printing
      process. Block ``a'' is the closed-loop position controller of the
      robot arm, while block ``b'' is the closed-loop position
      controller of the mobile base.}
    \label{fig:setup}
\end{figure}

The hardware setup is mostly the same as in our previous
work~\cite{zhang2018large}, to which the reader is referred for more
details. Briefly, the system comprises: a holonomic mobile base
(Clearpath Ridgeback), a 6-DoF industrial robot manipulator (Denso
VS-087) mounted on top of the mobile base, a nozzle mounted on the
manipulator flange and connected to a pump through a hose
(Fig.~\ref{fig:setup}, \textbf{Top}).

The main difference with~\cite{zhang2018large} is that the stereo
camera (Microsoft Kinect for Xbox One) is mounted on the back
of the mobile base, instead of being stationary. Conversely, the Aruco
markers~\cite{Garrido2014} of size 12\,cm$\times$12\,cm are placed on
the ground, instead of being placed on the robot as
in~\cite{zhang2018large}. This change enables the localization system
to be continuously effective in a larger area.

During execution, while the mobile base runs with its own battery, the
robot arm's power is supplied externally. Image processing, sensor
fusion and motion planning are performed on an off-board computer.


The pipeline of the process is depicted in Fig.~\ref{fig:setup},
\textbf{Bottom}. From the 3D model of the desired object, we first plan,
offline, the coordinated motions of the mobile base and the robot
arm in order to print the object layer by layer (Motion
planning). Next, we execute the planned motions on the actual
platform. During execution, the position of the mobile base is
monitored in real-time (Localization), and feedback control is
exercised to track the planned motions as close as possible (Motion
control). The following sections detail these processes.

\subsection{Motion planning}
\label{sec:motion_planning}

Compared to standard motion planning for a 6-DoF manipulator to cover a
large workspace (see e.g.~\cite{suarez2017}), the present motion
planning problem involves two additional difficulties
\begin{itemize}
\item The continuous printing process precludes switching between
  different Inverse Kinematics (IK) classes~\cite{xian2017closed};
\item The mobile base adds three extra (redundant) DoFs, as well as
  additional collision possibilities (between the robot and the base,
  or between the base and the printed specimen).
\end{itemize}

\begin{figure}[htp]
  \centering
  \includegraphics[width=0.9\columnwidth]{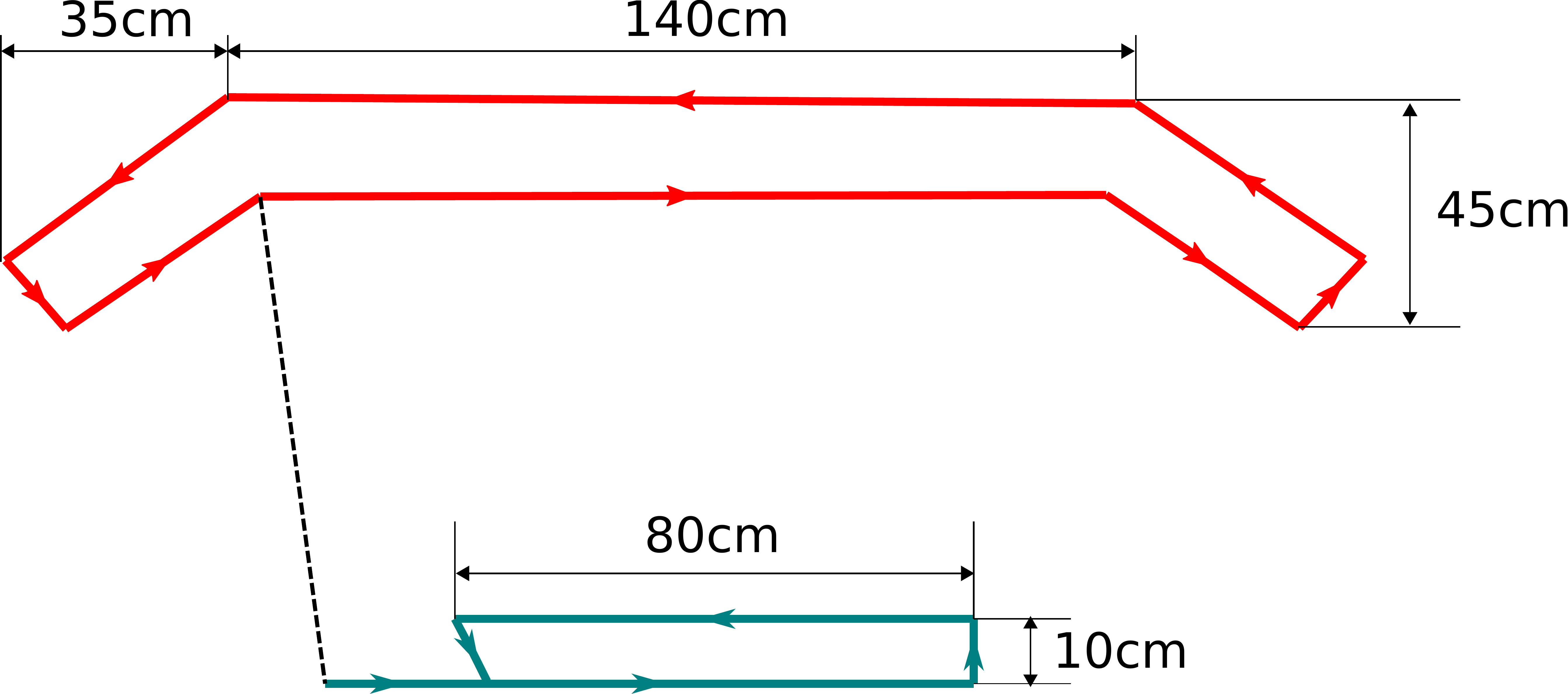} 
  \caption{Motion planning for the mobile base. Red: the nozzle path
    to print the specimen of Fig.~\ref{fig:robot}. Cyan: the path
    prescribed for the mobile base. The dashed black line depicts the
    synchronization of the two paths at the start of the printing.}
    \label{fig:path}
\end{figure}

To address these issues, we take the following simplified approach:
\begin{enumerate}
\item Prescribe a reasonable motion for the base, avoiding collision
  with the to-be-printed specimen and taking into account the
  manipulator reachability (Fig.~\ref{fig:path});
\item Synchronize the base motion with the nozzle motion (which
  follows the design of the specimen and has constant speed in order
  to deliver the printing materials at a constant rate);
\item Given the base motion and the nozzle motion, compute the joint
  trajectories for the manipulator by differential IK (see
  e.g.~\cite{xian2017closed}).
\end{enumerate}


Note that this approach is not optimal and is not guaranteed to always
work. For instance, the differential IK of step~3 might not find any
continuous motion for the manipulator, given the prescribed base and
nozzle motions. Developing a more principled planning algorithm for
mobile 3D Printing is part of our future work.

\subsection{Localization of the mobile base}
\label{sec:localization}

As stated in the Introduction, the accurate control of the mobile base
trajectory is crucial for printing quality. Note that, because the
accuracy of the industrial robot arm is significantly higher than that
of the mobile base, we make the assumption that the former is 100\%
accurate.

An essential component in trajectory control is localization (or pose
estimation), i.e. knowing accurately where the mobile base is at any
given time instant. Since the built-in localization module of the
mobile base has substantial drift over time, we use a vision-based
scheme: several fiducial Aruco markers are placed on the ground, and
when marker~$i$ is seen by the on-board camera, the position of the
mobile base can be estimated by
\[
  {}_{\mathcal{W}}\bfT^{\mathcal{B}}_{i} = {}_{\mathcal{W}}\bfT^{\mathcal{M}}_i
  {}_{\mathcal{M}}\bfT^{\mathcal{C}}_i {}_{\mathcal{C}}\bfT^{\mathcal{B}},\ \textrm{where}
\]
\begin{itemize}
\item $_{\mathcal{W}}\bfT^{\mathcal{B}}_{i}$ is the estimated homogeneous
transformation of the \underline{B}ase w.r.t. the \underline{W}orld
frame;
\item $_{\mathcal{W}}\bfT^{\mathcal{M}}_i$ is the (known) transformation
  of \underline{M}arker $i$ w.r.t. the World frame;
\item $_{\mathcal{M}}\bfT^{\mathcal{C}}_i$ is the transformation of the
  \underline{C}amera w.r.t. the marker (computed in real-time by the Aruco
  module~\cite{Garrido2014});
\item $_{\mathcal{C}}\bfT^{\mathcal{B}}$ is the (known) transformation of the
  Base w.r.t. the Camera.
\end{itemize}

From the transformation $_{\mathcal{W}}\bfT^{\mathcal{B}}_{i}$, one
can obtain the \underline{e}stimated position and orientation of the
mobile base $\bfx^e_i \in \mathrm{SE}(2)$. Next, to compensate for the
delay associated with image processing (which could be as long as
50\,ms), we propagate the estimated position and orientation using the
controls that were sent during the delay period by
\[
\bfx^p_{i,k} = \bfx^e_{i,k-N} + \Delta \sum_{j=k-N}^{k-1}\bfu_j,
\]
where $\bfx^p_{i,k}$ is the \underline{p}ropagated estimation at time
instant $k$, $\Delta$ is the control time step, $N$ is the number of
control time steps that have passed during the delay,
$\bfu_j := [v^x_j,v^y_j,\omega_j]$ is the control input at time step
$j$. Grouping together all the markers that are visible to the camera
at time step $k$, one finally obtain the measurement
\[
\breve{\bfx}_k := [\bfx^p_{1,k},\dots,\bfx^p_{M,k}],
\]
where $M$ is the number of visible markers at time step $k$. We make sure
that at least one marker is visible at any time instant.

Finally, we fuse the measurements with the controls using an Extended
Kalman Filter
\begin{itemize}
\item State prediction: $\bfx_{k+1} = \bfx_k+\Delta \bfu_k + \bfw^1_k$;
\item Measurement: $\bfz_k = \breve{\bfx}_k + \bfw^2_k$,
\end{itemize}
where $\bfw^1_k \sim \mathcal{N}(\mathbf{0},\bfQ)$ and
$\bfw^2_k \sim \mathcal{N}(\mathbf{0},\mathrm{diag}(\bfR,\dots,\bfR))$ are
respectively the prediction noise and the measurement noise at step
$k$. The values of the weight matrices $Q$ and $R$ were chosen
following a calibration process.

\subsection{Motion control for the mobile base}
\label{sec:motion_control}


There are two important design objectives in the control of the mobile
base: (i) track the desired trajectory accurately; (ii) obey the
velocity constraints for safe autonomous operation. In order to
address these objectives, we implement a Model Predictive
Controller. Note that the control inputs
$\bfu_k := [v^x_k,v^y_k,\omega_k]$ are calculated in the world frame.

The discrete dynamics of the mobile base for a horizon of $N$ steps is
given by
\begin{equation}
  \bfX_{k+1} = \underbrace{
    \begin{bmatrix}\bfA \\ \vdots\\ \vdots\\ \bfA^N 
    \end{bmatrix} }_{\bfS^x}
  \bfx_k+
  \underbrace{
    \begin{bmatrix}
      \bfB   & 0       & \ldots   & 0       \\
      \bfA\bfB   & \bfB & \ldots   & 0       \\
      \vdots     & \ddots  & \ddots   &  0      \\
      \bfA^N\bfB & \ldots  & \bfA\bfB & \bfB \\
    \end{bmatrix}}_{\bfS^u}
  \bfU_k,\nonumber
  \label{eqn:system_dynamics}
\end{equation}
where $\bfA:=\mathbb{I}_{3\times3}$,
$B:=\Delta \mathbb{I}_{3\times3}$,
$\bfX_{k+1} := [\bfx_{k+1}, \ldots, \bfx_{k+N}] \in
\mathbb{R}^{3N}$,
and
$\bfU_k := [\bfu_{k} \ldots \bfu_{k+N}] \in \mathbb{R}^{3N}$.

Next, consider the errors
\begin{eqnarray}
  \tilde{\bfx}_{k+i} &:=& \bfx_{k+i} - \bfx^d_{k+i},\nonumber \\
  \tilde{\bfu}_{k+i} &:=& \bfu_{k+i} - \bfu^d_{k+i},\nonumber
  \label{eqn:delta_variables}
\end{eqnarray}
where the superscript $d$ indicates the desired trajectory and
the nominal inputs. Note that $\bfu^d_{k+i}$ is zero in our case. The error
dynamics can then be written as
\[
\tilde{\bfX}_{k+1}=\bfS^x\bfx_{k}+ \bfS^u\bfU_{k}-\bar{\bfA} \bfX^d_{k+1},
\]
where
$\bar{\bfA}:=\mathrm{diag}(\bfA,\ldots,\bfA)\in\mathbb{R}^{3\times
  N}$, $\bfX^d_{k+1} := [\bfx^d_{k+1}, \ldots, \bfx^d_{k+N}] \in
\mathbb{R}^{3N}$.

The control inputs for horizon $N$ are calculated with the following
quadratic optimization 
\begin{align}
  \nonumber J^*(\bfx_k) = &\min_{\bfU} \quad \bfU^T \bfH\bfU + 2\bfq^T \bfU  \\
  \nonumber &\textrm{s.t.}  \quad \bfG\bfU \leq \bfw + \bfE \bfx_k +
              \bfE^d \bfX^d_{k+1},\\
  \nonumber            \bfH = &(\bfS^u)^T\bar{\bfQ}^c \bfS^u+\bar{\bfR}^c, \\
  \nonumber 
  \bfq = &(\bfx_k^T(\bfS^x)^T-(\bfX_{k+1}^d)^T\bar{\bfA}^T)\bar{\bfQ}^c\bfS^u.
\end{align}
where $\bar{\bfQ}^c:=\mathrm{diag}(\bfQ^c,\ldots,\bfQ^c)$ and
$\bar{\bfR}^c:=\mathrm{diag}(\bfR^c,\ldots,\bfR^c)$ are block-diagonal
weight matrices. The matrices $\bfG$, $\bfw$, $\bfE$ and $\bfE^d$
encode input and state constraints, see e.g.~\cite{WANG2009MPC} for
details.
 
\section{Experiments}
\label{sec:experiments}




\subsection{Accuracy and precision assessment}

\begin{figure}[htp]
  \centering
  \includegraphics[width=\columnwidth]{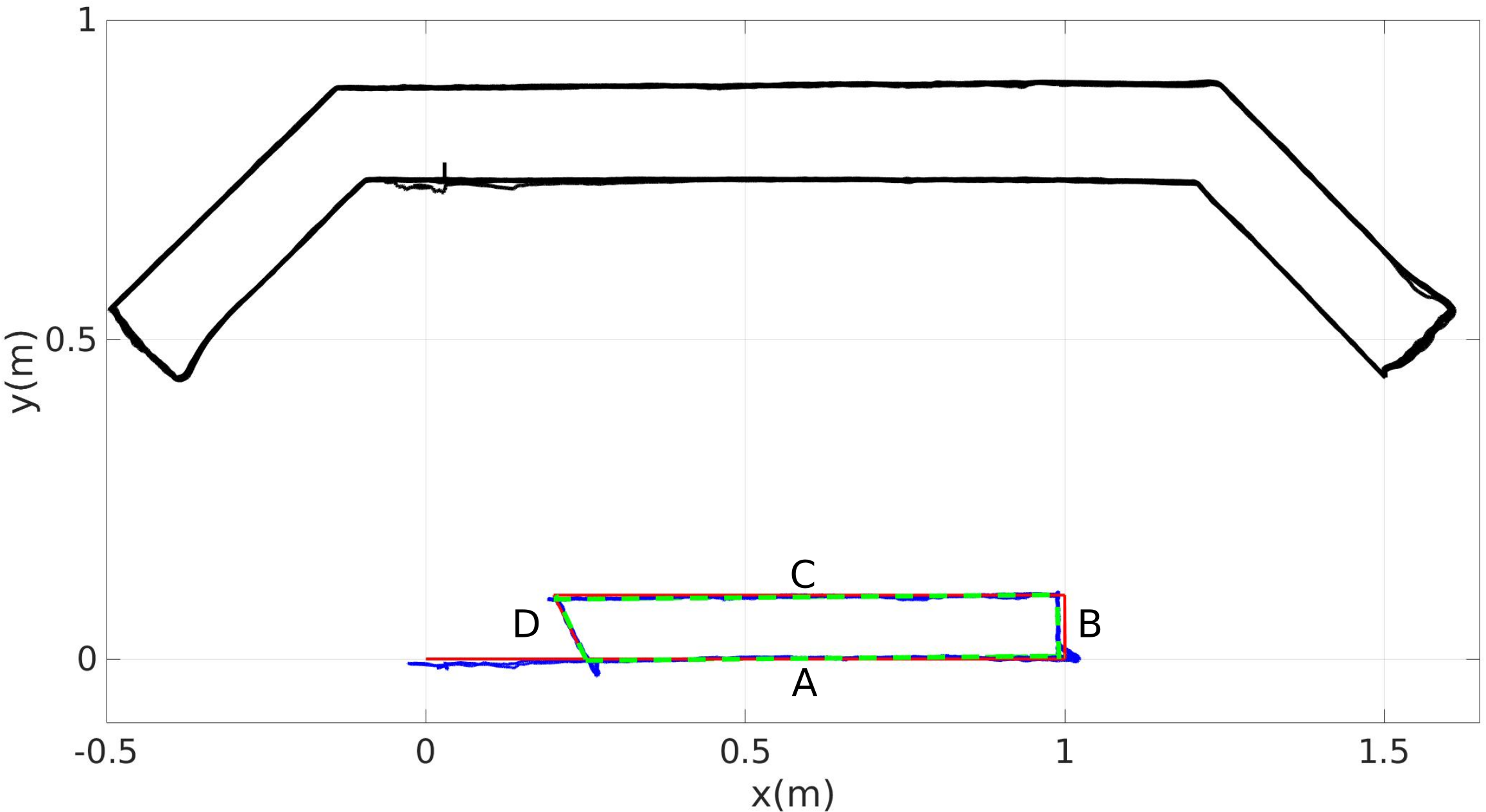}    
  \caption{Accuracy and precision assessment using an external
    Motion Capture (Mocap) system. Black: nozzle trajectory across
    15 laps as measured by Mocap. Blue: mobile base trajectory
    across 15 laps as measured by Mocap. Dashed green: straight
    lines fitted upon the blue lines. Red: desired trajectory for
    the base. The difference between the blue and dashed green lines
    reflects the precision, while the difference between the red and
    dashed green lines reflects the accuracy of our system.}
  \label{fig:mocap_plot}
\end{figure}

We used Optitrack, a Motion Capture system, with 6 cameras, to provide
an independent measurement. We placed markers on the nozzle and the
mobile base and capture their motions during an air printing
session. Fig.~\ref{fig:mocap_plot} shows the marker trajectories for
15 laps.

Before assessing the \emph{accuracy} of the system (how close the
printed shape is to the desired shape), we first evaluated its
\emph{precision} (how repeatable the system is), which is critical in
the layer-by-layer material deposition process. 

Consider a segment parallel to the X-axis (segments A and C in
Fig.~\ref{fig:mocap_plot}). We collected all the data points
$(x_i,y_i)_{i\in[1,N]}$ corresponding to this segment across the 15
laps. We then fit a line $y= ax + b$ to those data points. The
precision was then quantified by the maximum and the mean error
\begin{eqnarray}
  \nonumber e_{\textrm{max}} &:=& \max_i |y_i - (a x_i +b) |  \\
  \nonumber e_{\textrm{mean}} &:=& \sqrt{\frac{1}{N}\sum_i (y_i - (a x_i +b))^2}
\end{eqnarray}
For segments that are parallel to the Y-axis (segments B and D), we
swapped the roles of $x$ and $y$. Given the above definitions, the
maximum and mean errors across the four linear segments of base motion were respectively
9.9\,mm and 2.2\,mm.

Next, to assess the accuracy, we computed the maximum distance between
the fitted lines and the desired path, which was found to be 
9.8\,mm. 

Overall, the accuracy and precision of our system are significantly
better than those found in the SLAM literature. For example, a recent
review reports errors of more than 20\,mm in the best
case~\cite{Nardi2015}. Note however that our setup (use of fiducial
markers) and objectives are different from the SLAM literature. In any
case, the precision of our system was sufficient to print 10 layers of
concrete, as shown in the next section. Upon visual inspection of the
printed specimen, the print quality seems as good as in fixed-base
printing.

\subsection{Actual printing}

We tested the mobile printing system in the laboratory
environment. The structure to be printed is shown in
Fig.~\ref{fig:path} and has size a 210\,cm $\times$ 45\,cm $\times$
10\,cm, which is significantly larger than the reach of the robot arm
(87\,cm). We used a nozzle of diameter 1\,cm and a nozzle speed of
10\,cm/s. The cement was prepared as in the previous work of our
group~\cite{WENG2018,zhang2018large}.

\begin{figure*}[htp]
    \centering
    \includegraphics[width=\textwidth]{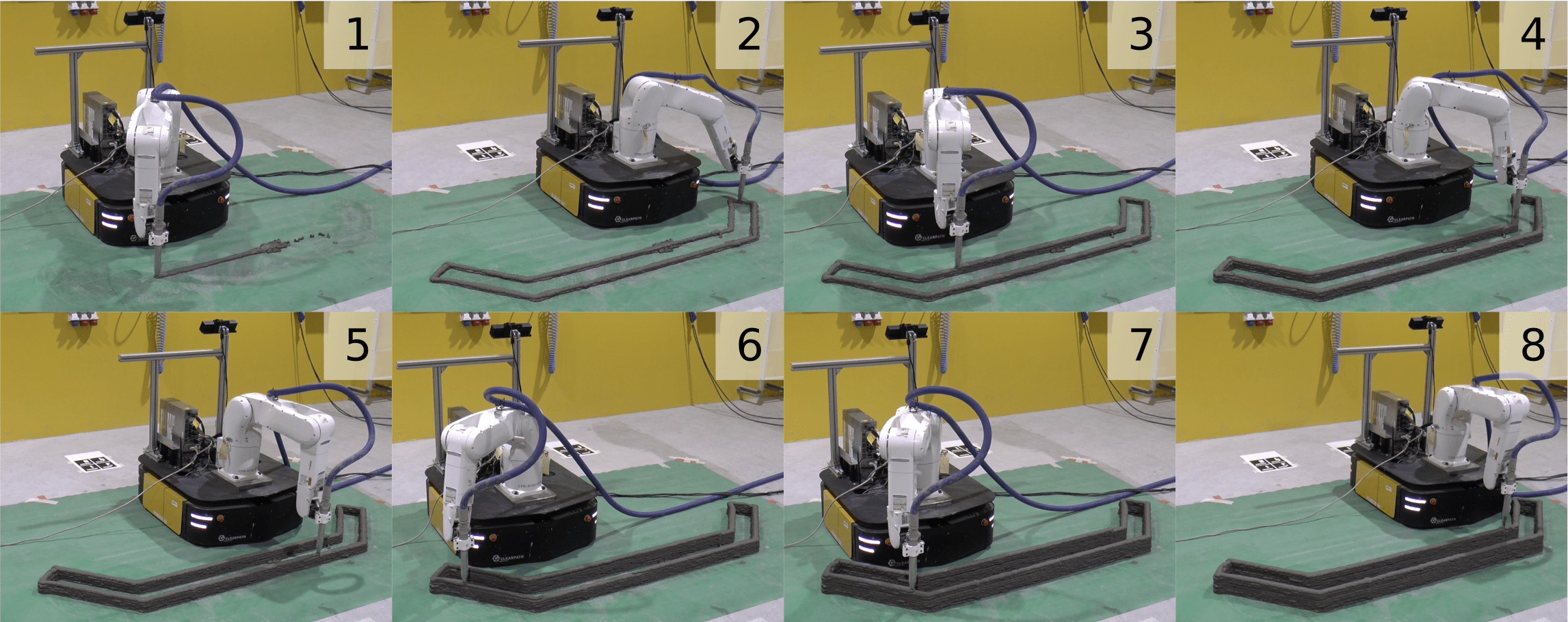}
    \caption{Snapshots of the actual printing experiment. A video of
      the experiment is available at
      \url{https://youtu.be/ZIbY00iTFwY}.} 
    \label{fig:time_sequence}
\end{figure*}

\begin{figure}[htp]
    \centering
    \includegraphics[width=\columnwidth]{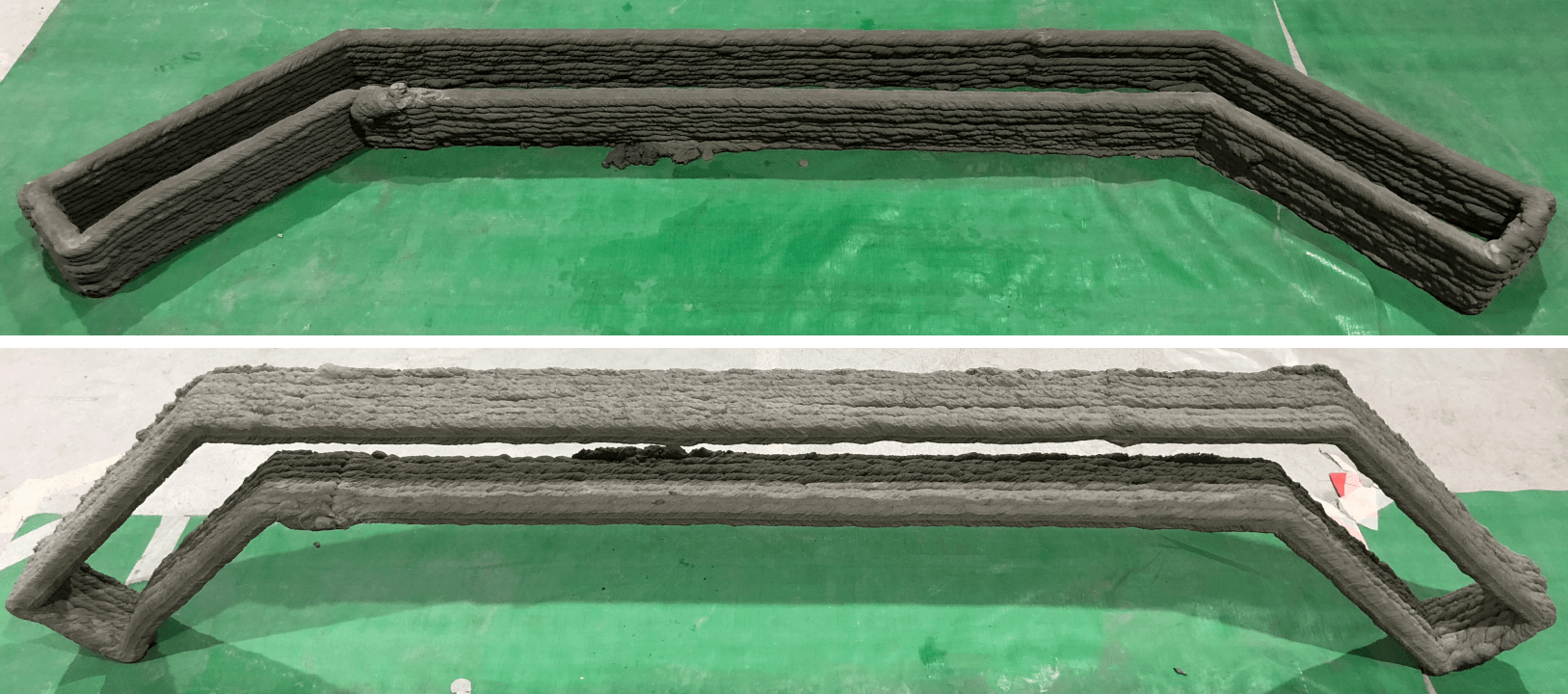}    
    \caption{Printed specimen after three days of curing.}
    \label{fig:specimen}
\end{figure}

We printed the 10 layers of the structure in 9\,min 16\,s. A video of
the experiment is available at
\url{https://youtu.be/ZIbY00iTFwY}. Snapshots of the printing session
are shown in Fig.~\ref{fig:time_sequence}. After three days of curing,
the structure was hard enough to be flipped and put on its side, as
shown in Fig.~\ref{fig:specimen}. As mentioned previously, the
precision of our trajectory tracking enabled to obtain a surface
finish similar to the specimen obtained by fixed-based
printing~\cite{zhang2018large}.


\section{Conclusion}
\label{sec:conclusion}

We have presented the printing-while-moving approach for large scale
in-situ concrete robotic 3D Printing. Our mobile printing system is
able to print single-piece concrete structures that are larger than
the robot reach, in a single take. We demonstrated, for the first
time, a printing-while-moving experiment, wherein a 210\,cm $\times$
45\,cm $\times$ 10\,cm concrete structure by a robot arm that has a
reach of 87\,cm.

Our system however still presents a number of
limitations. Localization errors may come from several sources: (i)
errors in the positions of the fiducial markers with respect to each
other and with respect to the world frame; (ii) vibrations of the
camera during the movements of the mobile base. Source (i) mainly
affects the accuracy, and can be mitigated by careful
calibration. Source (ii) can have a very large effect on the
precision, and can be mitigated by installing proper vibration
isolators for the camera. More generally, one can also integrate other
sensors, such as IMU or laser scanners, or implement more
sophisticated localization algorithms (e.g. based on particle
filters), to improve localization.

We observed that the sharp turns at the corners affected the precision
of our system in several manners. First, they cause vibrations of the
camera, which affects the vision-based localization as discussed
previously. Second, they cause vibrations of the nozzle, affecting the
material deposition. Third, the step-like control inputs are difficult
to track accurately by the physical system, causing large tracking
errors at the corners (see Fig.~\ref{fig:mocap_plot}). These issues
can be mitigated by planning smooth paths for the mobile base, or by
compensating the errors of the mobile base with fast arm motions.

Finally, the unevenness of the ground affects the quality of the
printing in two ways. First, at locations where the ground level rises
significantly, the extruded cement is squeezed more and spreads
laterally, resulting in a thinner layer in vertical direction. Second,
ground height variations also amplify the vibrations of the mobile
base. Again, those issues can be mitigated by mechanical vibration isolation or by
active disturbance rejection with the robotic arm.

Besides addressing the current limitations, there are also many
exciting avenues for further development. For example, the multi-robot
version of the presented concept can dramatically increases productivity,
but is intrinsically more challenging. It requires for instance a
motion planning algorithm that can plan optimal mobile base
trajectories considering multiple printing paths, multi-robot
collision avoidance, and material supply tethers.

\subsection*{Acknowledgement}

This work was supported in part by the Medium-Sized Centre funding
scheme (awarded by the National Research Foundation, Prime Minister’s
Office, Singapore) and by Sembcorp Design \& Construction Pte Ltd. We
would like to thank Lim Jian Hui, Weng Yi Wei, and Lu Bing for their
help with the experiment.

\balance
\bibliographystyle{IEEEtran}
\bibliography{bibliography/reference}

\end{document}